\documentclass[final,5p,times,twocolumn]{elsarticle}
\usepackage{amssymb}
\usepackage{amsmath}
\usepackage{url}
\usepackage{bbm}
\usepackage[caption=false]{subfig}
\usepackage{array,multirow,makecell}
\newcommand{\frameworkName}{Concept Directions via Latent Clustering}
\newcommand{\frameworkAcronym}{CDLC}

\newcolumntype{L}[1]{>{\raggedright\arraybackslash}p{#1}} 
\newcolumntype{C}[1]{>{\centering\arraybackslash}m{#1}}   

\newcommand{\supptitle}[1]{%
  \twocolumn[{\centering\Large\bfseries #1\par\vspace{1em}}]
}

\journal{Pattern Recognition Letters}

\begin{document}

\begin{frontmatter}

\title{Discovering Concept Directions from Diffusion-based Counterfactuals\\ via Latent Clustering}

\author[rptu,dfki]{Payal Varshney\corref{*}}
\ead{payal.varshney@dfki.de}

\author[rptu,dfki]{Adriano Lucieri}
\ead{adriano.lucieri@dfki.de}

\author[rptu,dfki]{Christoph Balada}
\ead{christoph.balada@dfki.de}

\author[rptu,dfki]{Andreas Dengel}
\ead{andreas.dengel@dfki.de}

\author[dfki]{Sheraz Ahmed}
\ead{sheraz.ahmed@dfki.de}

\cortext[*]{Corresponding author.}

\affiliation[rptu]{organization={Rheinland-Pfälzische Technische Universität Kaiserslautern-Landau},
            addressline={Gottlieb-Daimler-Street 47}, 
            city={Kaiserslautern},
            postcode={67663}, 
            state={Rhineland-Palatinate},
            country={Germany}}
\affiliation[dfki]{organization={German Research Center for Artificial Intelligence GmbH (DFKI)},
            addressline={Trippstadter Str 122}, 
            city={Kaiserslautern},
            postcode={67663}, 
            state={Rhineland-Palatinate},
            country={Germany}}

\begin{abstract}

Concept-based explanations have emerged as an effective approach within Explainable Artificial Intelligence, enabling interpretable insights by aligning model decisions with human-understandable concepts.
However, existing methods rely on computationally intensive procedures and struggle to efficiently capture complex, semantic concepts.
This work introduces the \frameworkName~(\frameworkAcronym), which extracts global, class-specific concept directions by clustering latent difference vectors derived from factual and diffusion-generated counterfactual image pairs.
\frameworkAcronym{} reduces storage requirements by $\sim$4.6$\times$ and accelerates concept discovery by $\sim$5.3$\times$ compared to the baseline method, while requiring no GPU for clustering, thereby enabling efficient extraction of multidimensional semantic concepts across latent dimensions.
This approach is validated on a real-world skin lesion dataset, demonstrating that the extracted concept directions align with clinically recognized dermoscopic features and, in some cases, reveal dataset-specific biases or unknown biomarkers.
These results highlight that \frameworkAcronym\ is interpretable, scalable, and applicable across high-stakes domains and diverse data modalities.

\end{abstract}


\end{frontmatter}


\section{Introduction}\label{sec:introduction}

In high-stakes applications, such as medical diagnosis, financial risk assessment, and autonomous driving, understanding the rationale behind a neural network’s decision is often as important as the decision itself.
Explainable Artificial Intelligence (XAI)~\cite{arrieta2020explainable, adadi2018peeking} has emerged as a critical research area, aiming to bridge the gap between high-performing black-box models and human interpretability.
Among the various XAI paradigms, concept-based explanations~\cite{kim2018interpretability, ghorbani2019towards} have gained particular attention due to their ability to express model behavior in terms of high-level, semantically meaningful concepts, rather than low‑level feature weights or pixel-based saliency maps~\cite {ribeiro2016should, selvaraju2017grad}.
By aligning explanations with concepts recognized by domain experts, these methods facilitate trust~\cite{shin2021effects, guo2020explainable}, debugging~\cite{das2021explainable}, and regulatory compliance~\cite{ebers2020regulating, panigutti2023role}.

Although concept-based explainability has been widely explored using convolutional~\cite{kim2018interpretability, ghorbani2019towards} and GAN-based architectures~\cite{lang2021explaining, atad2022chexplaining}, its application within diffusion-based generative models remains relatively underexplored.
Recent works, such as Concept-Guided Latent Diffusion Counterfactual Explanations (CoLa-DCE)~\cite{motzkus2024cola}, have demonstrated the ability to produce spatially constrained, concept-conditioned counterfactuals for an arbitrary classifier.
However, this approach only utilizes concepts to guide counterfactual explanations and does not extract global concept representations that are generalizable across examples.

Varshney et al.~\cite{varshney2025generating} proposed the Concept Discovery through Latent Diffusion-based Counterfactual Trajectories (CDCT) framework to discover global concepts that generalize across multiple samples.
CDCT generates a classifier-guided counterfactual image trajectory dataset using a latent diffusion model~\cite{rombach2022high} and subsequently trains a Variational Autoencoder (VAE)~\cite{kingma2013auto} on this trajectory dataset to disentangle classifier-relevant concepts.
While CDCT represents a significant advancement in leveraging diffusion-based counterfactuals for concept discovery, it relies on a dimension-wise traversal strategy, wherein each latent variable is modified independently to detect relevant concepts.
This exhaustive search procedure is computationally expensive, particularly in high-dimensional latent spaces, and inherently overlooks semantic concepts that arise from interactions among multiple latent dimensions.
Consequently, CDCT often identifies only simple concepts, restricting its capacity to uncover high-level semantic directions.

To address these limitations, this paper proposes a novel framework  \frameworkName~(\frameworkAcronym), that extracts multidimensional concepts by clustering latent difference vectors computed from factual–counterfactual image pairs.
These difference vectors, derived from VAE encodings of factual images and their corresponding diffusion-based counterfactuals, capture the classifier-induced transformation in latent space.
In contrast to CDCT, which modifies individual latent dimensions, \frameworkAcronym\ leverages directional clustering to reveal coordinated latent changes that correspond to semantically meaningful concepts.
This approach not only reduces computational complexity by eliminating exhaustive per-dimension search but also enables the discovery of classifier-relevant concept directions that emerge from interactions among multiple latent dimensions, effectively overcoming a limitation of the original CDCT formulation.

This paper makes the following key contributions: 

\begin{itemize}
    \item A novel framework, \frameworkName~(\frameworkAcronym), is introduced as an extension of CDCT~\cite{varshney2025generating}, significantly reducing computational complexity compared to dimension-wise search.

    \item \frameworkAcronym\ extracts global, multidimensional semantic concept directions by clustering latent-difference vectors obtained from VAE encodings of factual and diffusion-generated counterfactual images.
    
    \item The effectiveness of \frameworkAcronym\ is validated on a skin lesion classification task, where the discovered concept directions not only reliably flipped classifier predictions but also transferred robustly to unseen samples.  
\end{itemize}


\section{Related Work}\label{sec:relatedwork}
 
Explainable Artificial Intelligence (XAI) aims to make machine learning models more transparent by providing human-understandable insights into their decision-making processes~\cite{arrieta2020explainable, adadi2018peeking}.
Among various XAI approaches, concept-based explanations have gained increasing attention due to their ability to link internal model representations to semantically meaningful, human-interpretable concepts~\cite{kim2018interpretability, ghorbani2019towards, poeta2023concept}.
In supervised settings, concept-based methods rely on expert-provided annotations, which are costly and time-consuming to collect~\cite{kim2018interpretability, lucieri2020explaining}.
For instance, Testing with Concept Activation Vectors (TCAV)~\cite{kim2018interpretability} requires example sets for each concept and learns linear concept activation vectors to quantify a model’s sensitivity to those concepts. 
While effective, this reliance on curated concept examples limits the scalability and generalization of such methods.
To overcome these limitations, unsupervised concept-based methods have been introduced~\cite{ghorbani2019towards, fel2023craft}.
A prominent example is Automatic Concept Explanations (ACE)~\cite{ghorbani2019towards}, which discovers concepts by clustering segmented image patches, without requiring manual annotation.
However, ACE relies on spatial assumptions by considering localized image regions as potential concept candidates, limiting its ability to capture non-local or abstract concepts.
A comprehensive survey~\cite{poeta2023concept} has reviewed the landscape of concept-based explanations, outlining a range of methodologies and their respective applications.

Another prominent paradigm in XAI is counterfactual explanations, which provide “what-if” scenarios to reveal how minimal semantic changes to an input can alter a model’s prediction~\cite{wachter2017counterfactual}.
With the emergence of generative models, counterfactual synthesis has advanced significantly, generating more realistic and semantically meaningful examples.

In particular, diffusion-based approaches have been proposed to generate counterfactuals by incorporating classifier gradients into the denoising process~\cite{jeanneret2022diffusion,  jeanneret2023adversarial} or by applying adaptive parameterization and cone regularization of gradients~\cite{augustin2022diffusion}.
While these methods produce visually realistic counterfactuals, their explanations are typically local to individual samples.
In contrast, the Global Counterfactual Directions (GCD)~\cite{sobieski2024global} method learns latent directions that consistently invert sample classifications via a proxy model, enabling more generalizable interpretability.

Recent work has attempted to combine concepts with counterfactual generation.
CoLa‑DCE~\cite{motzkus2024cola} uses classifier guidance to produce spatially constrained, concept‑conditioned counterfactuals. 
Similarly, DiffEx~\cite{kazimi2025explaining} combines a vision–language model with diffusion semantics to generate a semantic hierarchy of attributes and rank their influence on classifier decisions. 
However, these methods remain sample-specific and do not yield reusable global concept vectors. 
In contrast, Varshney et al.~\cite{varshney2025generating} leveraged classifier-guided counterfactual trajectories to identify global disentangled semantic concepts.
Despite its strengths, CDCT’s dimension-wise traversal is computationally expensive and limited in detecting complex, multidimensional concepts.  

Another line of research leverages clustering in latent spaces to model semantic variation.
For example, DifCluE~\cite{jain2025difclue} clusters latent embeddings produced by a diffusion autoencoder to generate diverse counterfactual explanations, effectively modeling intra-class variation, but does not explicitly identify global concept directions.
A parallel line clusters concept-attribution vectors rather than latent features.
PCX~\cite{dreyer2024understanding} computes per-sample concept-relevance vectors and fits class-wise Gaussian mixture models to cluster them into prototypical decision strategies that explain model behavior.

However, most existing approaches reveal several open challenges: reliance on concept annotations or spatial heuristics, a focus on localized or sample-specific edits, the need for exhaustive latent traversal, and limited capacity to capture global, multidimensional semantic concepts.
To address these limitations, a novel framework, \frameworkName, is proposed as an extension of CDCT~\cite{varshney2025generating}. It discovers global, class-specific concept directions by clustering directional latent differences between factual and counterfactual images.
This approach enables the extraction of complex, multidimensional concept directions in an unsupervised manner and offers a computationally efficient framework for concept-based explanation, enhancing both interpretability and scalability.


\section{Methodology} \label{sec:methodology}

\begin{figure*}[h!]
    \centering
    \includegraphics[width=.95\linewidth]{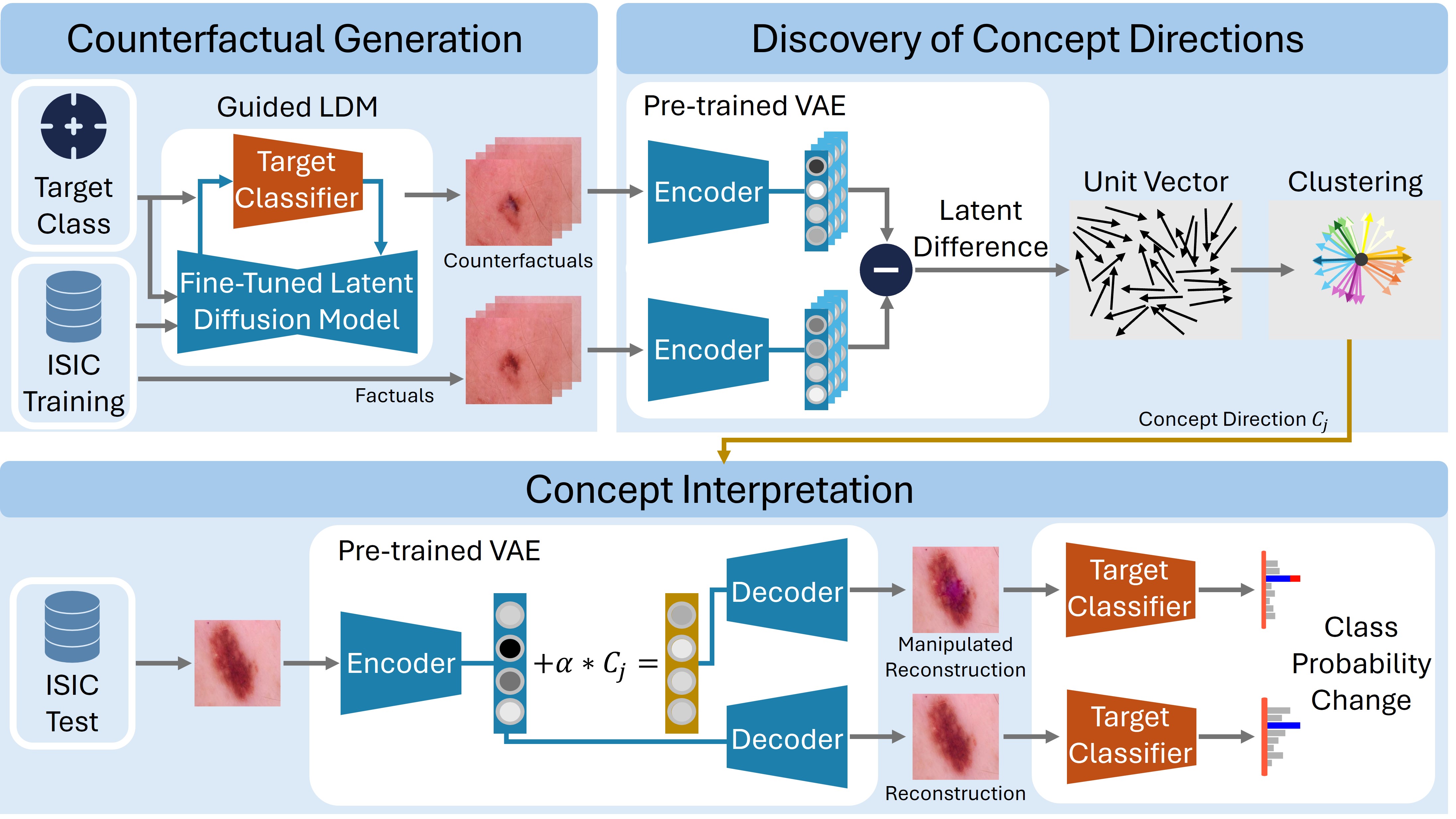}
    \caption{Overview of the \frameworkAcronym\ framework. Counterfactuals are generated using a Latent Diffusion Model (LDM) with classifier guidance, following the procedure used in CDCT~\cite{varshney2025generating}. Factual–counterfactual image pairs are encoded using a pretrained Variational Autoencoder (VAE), and the difference between their latent representations is normalized to form unit vectors. These vectors are clustered to identify class-specific concept directions $C_j$. During inference, each direction, scaled by a factor $\alpha$, is applied to the test sample's latent representation to observe its effect on classifier output.}
    \label{fig:Concept_discover_framework}
\end{figure*}

\frameworkName~(\frameworkAcronym) is a novel framework designed to extract global, multidimensional concept directions from factual and counterfactual image pairs.
It builds upon the counterfactual generation stage of CDCT~\cite{varshney2025generating}. However, instead of relying on computationally expensive dimension-wise latent traversals, it identifies concept directions by clustering unit-normalized difference vectors in a semantic latent space.
This section describes the key components of \frameworkAcronym\ in detail.

\subsection{Counterfactual Generation via Latent Diffusion Models}

Given an input image $x_f$ and a trained classifier $f(\cdot)$, a counterfactual image $x_{cf}$ is synthesized to belong to a target class $y_{cf} \ne f(x_f)$.
The counterfactual generation process follows the first stage of the CDCT framework~\cite{varshney2025generating}, termed \textit{Generation of Counterfactual Trajectories}, where classifier guidance is integrated into the denoising steps of a latent diffusion model~\cite{rombach2022high} to produce semantically meaningful modifications aligned with the desired class transition.
Details of this process can be found in the CDCT framework~\cite{varshney2025generating}.
Rather than capturing intermediate reconstructions across the trajectory, only the final counterfactual image is retained.
This design choice aims to capture the semantic shift required to alter the classifier's prediction while reducing storage and computational overhead. 
The resulting counterfactual is visually realistic and deviates minimally from the factual image, thereby isolating class-specific semantic changes.

\subsection{Discovery of Concept Directions}

Each factual image $x_f$ and its corresponding counterfactual $x_{cf}$ for a specific target class are encoded using a pretrained Variational Autoencoder, resulting in latent representations \( \mathbf{z}_f = \mathrm{VAE}(x_f) \) and \( \mathbf{z}_{cf} = \mathrm{VAE}(x_{cf}) \).
These latent embeddings capture high-level semantic characteristics of the original and counterfactual images in a compact form.

A difference vector is computed between the latent representations of each factual and counterfactual image pair as:
\[
\Delta \mathbf{z} = \mathbf{z}_{cf} - \mathbf{z}_f
\] 

Each difference vector is reshaped into a flat vector \( \Delta \mathbf{z} \in \mathbb{R}^d \).
To focus solely on the directional semantics and eliminate the influence of vector magnitude, each difference vector is normalized to unit length: 
\[
\tilde{\mathbf{z}} = \frac{\Delta \mathbf{z}}{\|\Delta \mathbf{z}\|_2}
\]

This transformation projects the latent differences onto the unit hypersphere in \( \mathbb{R}^d \), making them suitable for angular similarity-based clustering.

A collection of such unit-norm latent difference vectors \( \{\tilde{\mathbf{z}}_i\}_{i=1}^N \), derived from multiple factual–counterfactual pairs for a given target class, is clustered into $K$ clusters to identify shared semantic transformations.
Cluster centers \( \mathcal{C} = \{\mathbf{c}_1, \mathbf{c}_2, \ldots, \mathbf{c}_K\} \) are computed by averaging the unit vectors within each cluster and re-normalizing to unit length:
\[
\mathbf{c}_k \in \mathbb{R}^d, \quad \|\mathbf{c}_k\|_2 = 1
\]
Each cluster center \( \mathbf{c}_k \) represents a global concept direction, a consistent and interpretable shift in the latent space associated with the classifier's decision boundary.
These concept directions can then be applied to the latent representations of new test samples to induce semantically meaningful changes in the reconstructed output.

\subsection{Concept Interpretation}

To interpret each discovered concept direction \( \mathbf{c}_k \), it is applied to the latent encoding of an unseen test sample:

\[
\mathbf{z}'_{\text{test}} = \mathbf{z}_{\text{test}} + \alpha \cdot \mathbf{c}_k,
\]

where \( \alpha \in \mathbb{R} \) controls the strength of the semantic manipulation. The modified latent vector \(\mathbf{z}'_{\text{test}} \) is then passed through the VAE decoder to generate a concept-modified image. 

In contrast to CDCT, which iteratively modifies each latent dimension to search for influential concepts, \frameworkAcronym\ learns global, multidimensional concept directions directly from the difference of factual–counterfactual encodings.
\frameworkAcronym\ avoids storing intermediate image trajectories and relies on CPU-only clustering of latent differences; no auxiliary VAE training or iterative dimension-wise search is required.
These changes lead to a significant increase in computational efficiency and interpretable semantic transformations.

 
\section{Experiments \& Results}

This section presents the experimental setup and an analysis of the results produced by \frameworkName\ framework. 
Quantitative and qualitative evaluations assess the semantic coherence of the extracted concept directions and their effectiveness in influencing classifier predictions.

\subsection{Dataset and Classification Model}

The same dataset and classification architecture used in the CDCT framework~\cite{varshney2025generating} are adopted for evaluation. 
Experiments are conducted on a consolidated dermoscopic image dataset derived from the International Skin Imaging Collaboration (ISIC) challenges (2016–2020)\footnote{Dataset available at:~\url{https://challenge.isic-archive.com/challenges/}.}. 
Following the duplicate removal strategy proposed by Cassidy et al.~\cite{cassidy2022analysis}, a curated dataset of 29,468 unique images is obtained and stratified into training, validation, and test subsets.
The dataset includes eight diagnostic categories: \textit{Melanocytic Nevus} (NV), \textit{Melanoma} (MEL), \textit{Basal Cell Carcinoma} (BCC), \textit{Actinic Keratosis} (AK), \textit{Benign Keratosis} (BKL), \textit{Dermatofibroma} (DF), \textit{Vascular Lesions} (VASC), and \textit{Squamous Cell Carcinoma} (SCC).

A ResNet-50~\cite{he2016deep} model is trained on the ISIC training partition and used as the target classifier throughout all experiments. This network is used to generate classifier-guided counterfactuals and evaluate the effectiveness of discovered concept directions.

\subsection{Generation of Counterfactuals}

The same classifier-guided counterfactual generation approach introduced in the CDCT framework~\cite{varshney2025generating} is adopted, with a simplified setting.
Rather than capturing intermediate reconstructions across the trajectory, only the final counterfactual image is retained.
Counterfactuals are generated using a Latent Diffusion Model based on the Stable Diffusion (SD) 2.1 architecture\footnote{Model available at:~\url{https://huggingface.co/stabilityai/stable-diffusion-2-1}.}, which is fine-tuned on the consolidated ISIC training dataset to align the generative manifold with the dermoscopic image domain.
For counterfactual generation, the same hyperparameters (guidance scale, diffusion steps) are adopted as reported in CDCT, to ensure consistency and comparability (see Appendix A for hyperparameters).
For each training sample, counterfactuals are synthesized for all target classes other than the class predicted by the classifier. 

\subsection{Extraction of Concept Directions}

Semantic concept directions induced by classifier-guided counterfactuals are explored in two distinct latent spaces: 
(1) the pretrained encoder of the Stable Diffusion 2.1 model, referred to as the \textit{LDM encoder}, and 
(2) a Variational Autoencoder trained on counterfactual trajectories, referred to as the \textit{CDCT encoder} (see Step~2 of the CDCT framework~\cite{varshney2025generating} for further details).
Experimental setup, results, and analysis based on the CDCT encoder are provided in the Supplementary Material (Appendix E, F).

To enable class-specific concept discovery, latent difference vectors are computed separately for each target class.
For this purpose, training samples not predicted as the target class are selected, and counterfactual images are generated that shift the classifier's prediction to the target class.
The corresponding factual and counterfactual images are encoded using the LDM encoder, resulting in latent embeddings of shape $4 \times 32 \times 32$.
The element-wise difference between these embeddings captures the transformation required to alter the prediction.
Each difference tensor is flattened into a $4096$-dimensional vector and normalized to unit length, forming a directional vector in latent space.

The unit vectors are collected for all training samples that are not predicted as a given target class.
Spherical K-Means clustering~\cite{hornik2012spherical} is then applied to group these vectors that reflect similar semantic transformations.
The number of clusters $K$ is selected based on the highest silhouette score (see Appendix A for per-class $K$ values).
A detailed ablation study analyzing the effect of the number of clusters $K$ is provided in the Supplementary Material (Appendix B).
Each resulting cluster represents a set of consistent latent transformations, and the average unit direction within a cluster is interpreted as a representative concept direction for the target class.

\subsection{Interpretation and Evaluation of Concept Directions}

\begin{figure*}[htp]
    \centering  
    \includegraphics[width=.90\textwidth]{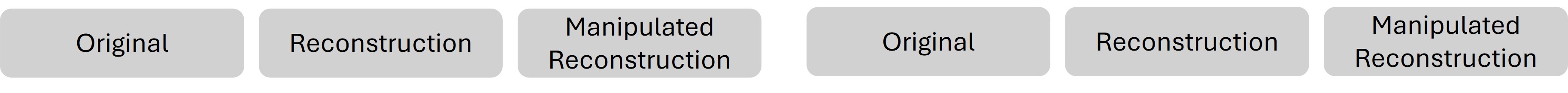} \\ 
    \centering
    \subfloat[Success Rate = 70.54 \% - Purplish Core Pigmentation.]{\label{fig:MEL_LDM}
    \includegraphics[width=.90\textwidth]{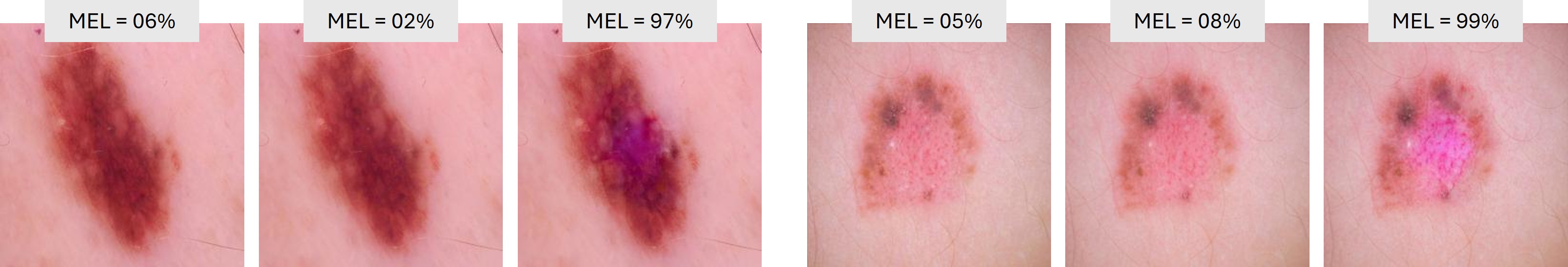}} \\    
    \subfloat[Success Rate = 83.61 \% - Reddish Core with Blue-Gray Dots.]{\label{fig:BCC_LDM}
      \includegraphics[width=.90\textwidth]{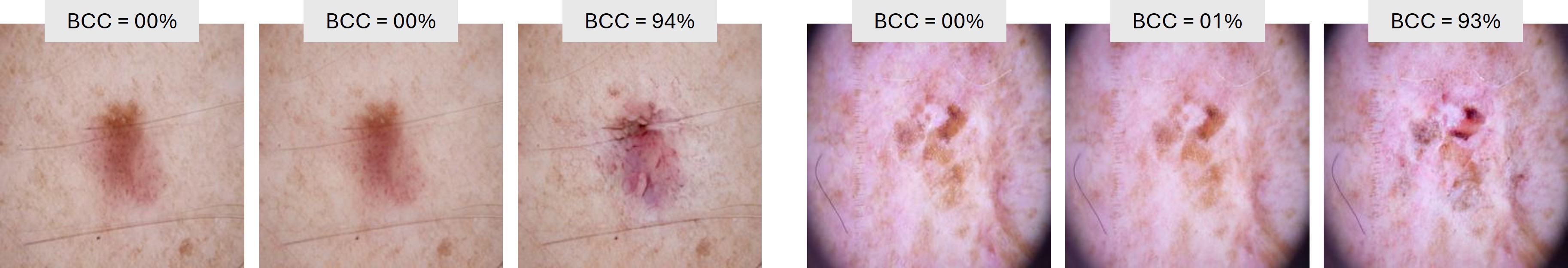}} \\      
    \subfloat[Success Rate = 81.09 \% - Blotchy Pigmentation with Irregular Texture.]{\label{fig:BKL_LDM}
      \includegraphics[width=.90\textwidth]{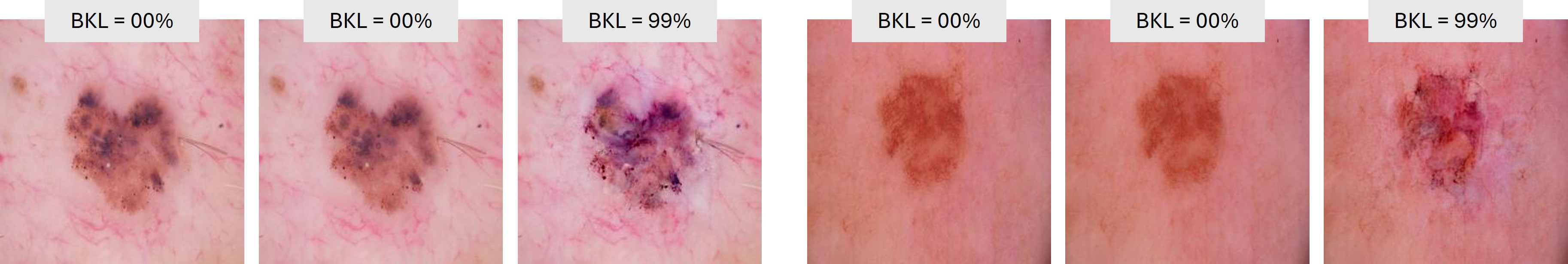}} \\      
    \subfloat[Success Rate = 85.44 \% - Central Pinkish Veil with Asymmetric Intensified Pigmentation.]{\label{fig:DF_LDM}
      \includegraphics[width=.90\textwidth]{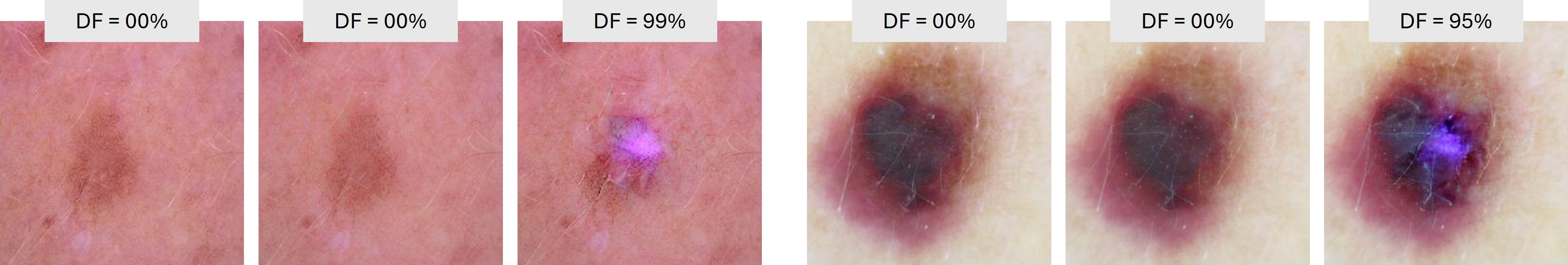}} \\      
    \subfloat[Success Rate = 84.59 \% - Central Purplish Veil.]{\label{fig:VASC_LDM}
      \includegraphics[width=.90\textwidth]{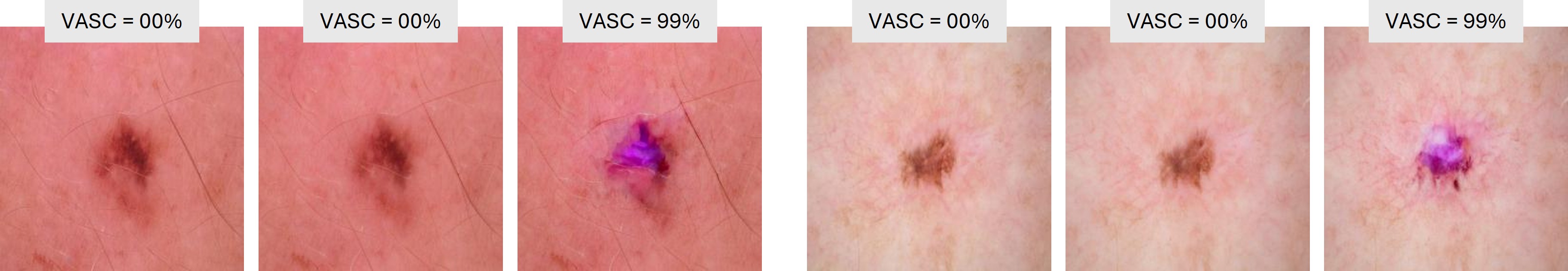}} \\      
    \subfloat[Success Rate = 70.47 \% - White Structures with Irregular Vessels.]{\label{fig:SCC_LDM}
      \includegraphics[width=.90\textwidth]{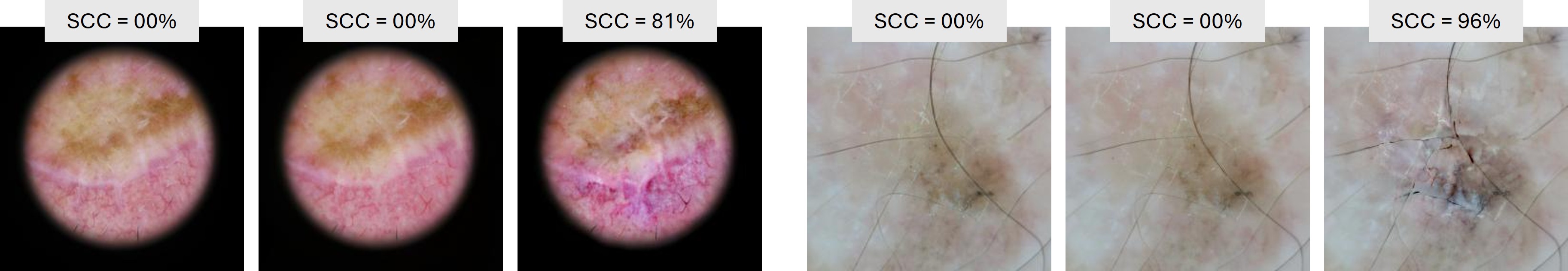}}      
    \caption{Discovered concepts by \frameworkAcronym\ on the ISIC dataset using the LDM encoder. Each row shows two examples: original, reconstructed, and manipulated reconstruction (left to right). The predicted probability for the target class associated with each concept direction is shown above each image.}
    \label{fig:concepts_LDM}
\end{figure*}

To assess the interpretability and effectiveness of the extracted concept directions, each identified concept direction is applied to unseen test images in the latent space. 
A detailed ablation study analyzing the effect of the scaling parameter $\alpha$ on concept traversal is provided in the Supplementary Material (Appendix C).
Figure~\ref{fig:concepts_LDM} illustrates concept directions identified by \frameworkAcronym\ using the LDM encoder.
For \textit{Melanoma}, the extracted direction (Figure~\ref{fig:MEL_LDM}) increases central pigmentation and introduces a purplish hue in the core lesion. 
In \textit{Basal Cell Carcinoma}, detected concept direction (Figure~\ref{fig:BCC_LDM}) produces a reddish-purple hue with blue-gray dots.
For \textit{Benign Keratosis}, a concept direction (Figure~\ref{fig:BKL_LDM}) induces purplish blotches, irregular pigmentation, and blurred borders.
In the case of \textit{Dermatofibroma}, the direction (Figure~\ref{fig:DF_LDM}) introduces a violet-toned center with subtle border irregularities and increased asymmetric pigmentation.
For \textit{Vascular Lesions}, the direction (Figure~\ref{fig:VASC_LDM}) enhances central coloration with purplish tones and peripheral pink diffusion.
Finally, \textit{Squamous Cell Carcinoma} directions (Figure~\ref{fig:SCC_LDM}) reveal white central structures accompanied by irregular linear streaks suggestive of vascular features.

Table~\ref{tab:quan_cdlc} reports the quantitative results for the detected concept directions.
Success Rate (SR) measures the proportion of test samples where traversal along the direction increases the probability of the target class, relative to the reconstructed image.
LPIPS~\cite{zhang2018unreasonable} and FID~\cite{heusel2017gans} assess perceptual and distributional realism, while TCAV quantifies concept alignment with the model’s decision boundary, computed from the final conv3 outputs of the third bottleneck block in layer 4 of ResNet50. 
A detailed analysis of TCAV computed across different layers of ResNet50 for these concept directions can be found in the Supplementary Material (Appendix D).
Most concepts achieve high SR (70–85\%), confirming their strong influence on classifier decisions. 
Concepts such as \textit{Central Pinkish Veil} and \textit{Central Pinkish Veil with Asymmetric Intensified Pigmentation} attain both high SR and near-perfect TCAV, indicating consistent use by the model.
In contrast, \textit{White Structures with Irregular Vessels} show lower SR and TCAV, suggesting weaker alignment. 
Overall, SR, LPIPS, FID, and TCAV demonstrate that the discovered concepts are semantically meaningful and predictive of model behavior.

\setlength{\arrayrulewidth}{1.2pt}
\begin{table}[!h]
\setlength{\tabcolsep}{2.4pt}
\caption{\label{tab:quan_cdlc} Quantitative results for the detected concept directions, reported in terms of Success Rate (SR), LPIPS, FID, and TCAV. Together, these metrics demonstrate that the discovered concepts are both semantically meaningful and predictive of model behavior.}
\centering
\begin{tabular}{>{\raggedright\arraybackslash}m{3.45cm}cccc}
\hline
\textbf{Concept} & \textbf{SR (\%)}~$\uparrow$ & \textbf{LPIPS}~$\downarrow$ & \textbf{FID}~$\downarrow$ & \textbf{TCAV}~$\uparrow$\\
\hline
\makecell[l]{Purplish Core\\ Pigmentation} & 70.5 & 0.12 & 30.5 & 0.97 \\
\hline
\makecell[l]{Reddish Core with \\ Blue-Gray Dots} & 83.6 & 0.17 & 43.1 & 0.82 \\
\hline
\makecell[l]{Blotchy Pigmentation\\ with Irregular Texture} & 81.1 & 0.20 & 47.0 & 0.92 \\
\hline
\makecell[l]{Central Pinkish Veil\\ with Asymmetric\\ Intensified Pigmentation} & 85.4 & 0.15 & 47.1 & 1.00 \\
\hline
\makecell[l]{Central Purplish Veil} & 84.6 & 0.15 & 53.5 & 1.00 \\
\hline
\makecell[l]{White Structures\\ with Irregular Vessels} & 70.5 & 0.18 & 51.2 & 0.64 \\
\hline
\end{tabular}
\end{table}

\subsection{Computational Efficiency}

To assess computational efficiency, \frameworkAcronym{} is compared with CDCT for detecting concept directions of the \textit{Melanoma} class.
As shown in Table~\ref{tab:runtime_storage}, both methods have comparable generation times, but \frameworkAcronym{} requires $\sim$4.6$\times$ less storage by retaining only latent differences instead of entire trajectories. 
Beyond storage, CDCT requires additional architectural overhead, including training an auxiliary VAE and performing iterative dimension-wise traversal. In contrast, \frameworkAcronym{} requires no additional training; concept discovery reduces to clustering latent differences, which runs efficiently on CPU. 
Despite running on CPU, \frameworkAcronym{} represents a $\sim$5.3$\times$ speedup over CDCT’s GPU-based extraction step. 
Overall, \frameworkAcronym{} substantially reduces computation time and storage, offering a scalable and practical alternative without sacrificing interpretability.

\begin{table}[h]
\centering
\caption{Runtime and storage comparison between \frameworkAcronym{} and CDCT for detecting concept directions of the \textit{Melanoma} class. All GPU operations were performed on an NVIDIA L40S GPU.  \frameworkAcronym{} achieves $\sim$4.6$\times$ lower storage requirements and $\sim$5.3$\times$ faster concept extraction despite running on CPU.}

\begin{tabular}{lcccc}
\hline
\textbf{Method} & \textbf{Hardware} & \textbf{Time} & \textbf{Storage} \\
\hline
CDCT (trajectory) & GPU & 1d 16h 25m & 314 MB \\
\frameworkAcronym\ (embed. diff.) & GPU & 1d 18h 35m & 69 MB \\
CDCT (dim. search) & GPU & $\sim$80 m & --- \\
\frameworkAcronym\ (clustering) & CPU & $\sim$15 m & --- \\
\hline
\end{tabular}
\label{tab:runtime_storage}
\end{table}


\section{Discussion}

The results demonstrate that \frameworkAcronym\ effectively extracts global concept directions for skin lesion classification while significantly improving computational efficiency over CDCT.
By avoiding iterative traversal of individual latent dimensions, \frameworkAcronym\ reduces the overhead typically associated with concept discovery, especially in high-dimensional latent spaces.
\frameworkAcronym{} achieves a $\sim$4.6$\times$ reduction in storage requirements and a $\sim$5.3$\times$ speedup in concept extraction relative to CDCT, and clustering in \frameworkAcronym{} executes entirely on CPU without requiring GPU resources.
Additionally, it captures rich, multidimensional semantic directions beyond the axis-aligned perturbations of CDCT, which tend to detect only simple features.
Beyond efficiency, \frameworkAcronym\ reliably discovers class-specific concept directions by constructing latent difference vectors separately for each target class.
This change avoids feature entanglement observed in CDCT, where a single VAE is trained on trajectories across all classes.
As a result, the extracted directions are more semantically coherent and clinically interpretable.

To further analyze the visual characteristics of the extracted directions, the behavior of the two latent encoders used in the framework is compared. 
While both encoders support meaningful concept discovery, the CDCT encoder often fails to reconstruct subtle yet clinically relevant features, limiting the fidelity of concept manipulations.
In contrast, the LDM encoder preserves fine-grained details and structural consistency, yielding more realistic and diagnostically coherent manipulations.
Nonetheless, directions in the LDM space frequently introduce pinkish or violet hues, which may influence the visual appearance of concepts. Possible explanations include the fact that \frameworkAcronym\ learns directions in latent difference space and does not explicitly optimize for hue variations; such hue shifts may therefore arise from generative priors in the latent-diffusion model, from color correlations learned by the target classifier, or from relevant color biomarkers. Therefore, color changes alone are not interpreted as concepts; instead, the qualitative analysis emphasizes structural and textural cues that consistently co-occur with the discovered directions.

The concept directions discovered by \frameworkAcronym\ demonstrate alignment with established dermoscopic features.
For instance, directions targeting \textit{Melanoma} often exhibit asymmetric pigmentation and color variegation, features of malignancy reported in clinical literature~\cite{argenziano2003dermoscopy}.
In \textit{Basal Cell Carcinoma}, the extracted directions introduce translucent reddish or violet hues with dark pigment patches, consistent with features such as ulceration and blue-gray dots~\cite{altamura2010dermatoscopy}.
For \textit{Dermatofibroma}, directions emphasize a central scar-like area of pink hues reported in literature~\cite{agero2006conventional}, while \textit{Vascular Lesions} display purplish centers with surrounding pink diffusion, resembling the vascular blush of angiomas~\cite{piccolo2018dermatoscopy}.
Similarly, directions for \textit{Squamous Cell Carcinoma} reveal white scaly textures and irregular vascular patterns, characteristic of keratinizing and sun-damaged lesions~\cite{rosendahl2012dermoscopy}.

In addition to replicating known diagnostic cues, \frameworkAcronym\ also uncovers subtler variations such as peripheral pigmentation or central hue changes, which may indicate dataset-specific biases or underexplored clinical markers, and requires clinical feedback.
In contrast, reliable concept directions could not be identified for the \textit{Nevus} class. This is likely due to class imbalance, as the dominance of \textit{Nevus} in the training data results in fewer samples being mapped as counterfactual to it. Consequently, the limited number of latent difference vectors hinders the discovery of consistent semantic directions for this class.


\section{Limitations \& Future work}

While \frameworkAcronym\ demonstrates promising results in extracting semantically meaningful and class-specific concept directions, several limitations remain.
The reliability of extracted directions depends on the quality of counterfactuals, as artifacts or semantic drift at this stage can introduce noise into the latent differences.
Some concept directions may exhibit hue shifts (e.g., pink/violet), potentially reflecting generative priors, color correlations in the classifier, or potential color biomarkers. Disambiguation of these causes would require targeted controls (e.g., color-invariant rendering, grayscale analyses) and clinical validation, which we consider beyond the scope of this work and leave for future work.
Additionally, spherical clustering assumes that concept directions are well-separated based on angular similarity on a unit hypersphere, which may not adequately capture complex latent geometries or overlapping semantic structures. Exploring more flexible clustering approaches (e.g., kernelized or manifold-based methods) remains an avenue for future work.

Future work could explore alternative encoders, such as Diffusion Autoencoders~\cite{preechakul2022diffusion}, to enhance reconstruction fidelity and semantic expressiveness.
Extending \frameworkAcronym{} to other modalities, validating its transferability across domains, and testing it with alternative architectures may further broaden its applicability.
Finally, incorporating human-in-the-loop evaluation or clinical feedback would be essential to establish the medical validity of the discovered concepts and support their real-world integration. Such validation requires substantial time and collaboration with domain specialists and is left for future work.

 
\section{Conclusion}

This work introduced \frameworkAcronym, a framework for discovering global, class-specific concept directions by clustering latent differences between factual and counterfactual image pairs.
Experiments on a skin lesion classification task demonstrated that the extracted directions influence model predictions and align with known dermoscopic features, supporting their interpretability.
However, some concept directions, especially obtained from the LDM encoder, consistently exhibit pinkish or purplish hues, irrespective of the target class.
This visual redundancy may indicate an inherent bias in the generative prior, suggesting a need for further analysis to disentangle meaningful concepts from model or dataset-specific artifacts.
The ability to uncover diverse, clinically relevant concepts positions \frameworkAcronym\ as a valuable tool for advancing concept-based explainability in high-stakes domains.

\section{Acknowledgments}
The project was funded by the Federal Ministry for Education and Research (BMBF) with grant number 03ZU1202JA.

\section{CRediT authorship contribution statement}
\textbf{Payal Varshney:} Conceptualization, Data curation, Formal analysis, Investigation, Methodology, Software, Validation, Visualization, Writing – original draft.
\textbf{Adriano Lucieri:} Conceptualization, Data curation, Supervision, Validation, Writing – review and editing.
\textbf{Christoph Balada:} Writing – review and editing.
\textbf{Andreas Dengel:} Funding acquisition, Project administration, Resources, Supervision, Writing – review and editing.
\textbf{Sheraz Ahmed:} Supervision, Writing – review and editing.

\section{Declaration of generative AI and AI-assisted technologies in the writing process}
During the preparation of this work the authors used GPT-4o in order to improve language and readability. After using this tool/service, the authors reviewed and edited the content as needed and take full responsibility for the content of the publication.

\appendix
\supptitle{Supplementary Material}

\section{Hyperparameters}

For counterfactual generation, we followed the CDCT~\cite{varshney2025generating} framework and adopted the same hyperparameter settings: guidance scale = 4 and number of diffusion steps = 10. 
The number of clusters $K$ was determined using the highest silhouette score. 
The selected $K$ values for each class are reported in Table~\ref{tab:clusters}. 

\setlength{\arrayrulewidth}{1.2pt}
\setlength{\tabcolsep}{4pt}
\begin{table}[h]
\centering
\renewcommand{\arraystretch}{1.3}
\caption{Number of clusters $K$ selected for each skin lesion class using highest the silhouette score. \\}
\begin{tabular}{lc}
\hline
\textbf{Class} & \textbf{Clusters $K$} \\
\hline
Melanoma                & 8 \\
Nevus                   & 5 \\
Basal Cell Carcinoma    & 6 \\
Benign Keratosis        & 7 \\
Actinic Keratosis       & 6 \\
Dermatofibroma          & 7 \\
Vascular Lesions        & 6 \\
Squamous Cell Carcinoma & 5 \\
\hline
\end{tabular}
\label{tab:clusters}
\end{table}

\section{Ablation Study on the Number of Clusters $K$}

For each lesion class \(y\), the number of clusters \(K\) is determined using the silhouette score.  
To validate this choice, an ablation study is performed by varying \(K\in\{4,5,6,7,8,9\}\) and measuring the downstream effects and the diversity of the learned concept directions.  

Let \(X\) denote the test set.  
For class \(y\) and image \(x\in X\), the \(k\)-th concept direction (cluster center) is denoted by \(c_{y,k}\), and the model probability for class \(y\) by \(f_y(\cdot)\).  
Applying a concept with scaling \(\alpha\) produces a transformed input
\[
x_{\alpha,k} \;=\; T(x;\,\alpha,\,c_{y,k}),
\]
and the corresponding per–concept effect is defined as
\[
\Delta_y(x,k) \;=\; f_y(x_{\alpha,k}) \;-\; f_y(x).
\]

\noindent\textbf{Coverage.}
Coverage measures the fraction of images for which at least one concept produces a positive effect above a small threshold \(\delta>0\) (e.g., \(\delta=0.05\)):
\[
\mathrm{Coverage}_y(K) \;=\; \frac{1}{|X|} \sum_{x\in X} 
\mathbbm{1}\!\left[\max_{1\le k\le K}\Delta_y(x,k)\,\ge\,\delta\right].
\]
This metric captures the availability of at least one relevant concept for a given image.\\

\noindent\textbf{Best-of-$K$ influence.}
The Best-of-\(K\) influence measures the gain per image when the most relevant concept is applied:
\[
\mathrm{BestInfluence}_y(K) \;=\; \frac{1}{|X|} \sum_{x\in X} \max_{1 \le k \le K} \Delta_y(x,k).
\]
This metric serves as an upper bound on actionable improvement, without requiring any single concept to be effective for all images. \\

\noindent\textbf{Robust mean influence (top-$q$ mean).}
A robust average is computed to reduce the impact of weak concepts.
For \(q\in(0,1]\) (with $q=0.3$ in our experiments) and \(m=\lceil qK\rceil\), the top-\(q\) mean is
\[
\mathrm{TopQMean}_y(K) \;=\; \frac{1}{|X|} \sum_{x\in X} \frac{1}{m} \sum_{k\in \mathrm{top}_m\{\Delta_y(x,1{:}K)\}}
\Delta_y(x,k).
\]
This measure captures the contribution of several useful concepts per image while limiting the influence of outliers. \\

\noindent\textbf{Redundancy index (diversity).}
A redundancy index quantifies diversity among concept directions.  
With unit–norm directions \(\hat c_{y,k}\) for \(k=1,\dots,K\), redundancy is defined as the mean pairwise cosine similarity between all distinct pairs of directions:
\[
\mathrm{Redundancy}_y(K) \;=\; \frac{2}{K(K-1)} \sum_{1 \le i < j \le K} \big\langle \hat c_{y,i},\,\hat c_{y,j} \big\rangle.
\]
Lower values indicate more diverse (less duplicated) concepts, while orthogonal sets yield values near zero. \\

Table~\ref{tab:ablation_k} reports these metrics for \(K\in\{4,5,6,7,8,9\}\) for each class \(y\). 
Across classes, Coverage, Best-of-\(K\), and the top-\(q\) mean vary only slightly with \(K\), indicating that downstream performance is largely stable with respect to the number of clusters. 
Best-of-\(K\) tends to increase as \(K\) grows, while Coverage and the top-\(q\) mean show only small, class-specific gains. 
Redundancy exhibits modest class-specific fluctuations and is often lowest near the silhouette-selected \(K\). 
Although the exact concept directions differ when \(K\) changes, their aggregate usefulness and diversity remain stable. 
This study confirms that the silhouette-based choices reported in the main text (Mel=8, BCC=6, BKL=7, DF=7, VL=6, SCC=5) represent robust operating points: they achieve near-maximal performance while avoiding unnecessary redundancy, and the results are not sensitive to the precise choice of \(K\). 
Moreover, this ablation framework provides a general strategy for selecting an appropriate number of clusters. 
When the method is transferred to other domains, the same analysis can guide the choice of \(K\) in a robust and data-driven manner.

\begin{table*}[!h]
\caption{\label{tab:ablation_k} Ablation study on the number of clusters \(K\) for each lesion class \(y\). Reported metrics: Redundancy Index (lower is better), Coverage, Best-of-\(K\), and Robust Mean Influence (top-\(q\) mean, \(q{=}0.3\)). Boldface in the \(K\) column marks the silhouette-selected \(K\); boldface within metric columns marks per-class best values.\\} 
\centering 
\renewcommand{\arraystretch}{1.3}
\begin{tabular}{L{2.5cm} C{2cm} C{2cm} C{2cm} C{2cm} C{2.2cm}} 
\hline
\textbf{Target Class} & \textbf{Number of Clusters $K$} & \textbf{Redundancy Index} & \textbf{Coverage} & \textbf{Best-of-K Influence } & \textbf{Robust Mean Influence (top-q mean)} \\
\hline
\multirow{6}{2.5cm}{ Melanoma}  & 4          & 0.69 & 0.72 & 16.7 & 13.7 \\
                                & 5          & 0.72 & 0.75 & 19.1 & 15.8 \\
                                & 6          & 0.73 & 0.76 & 20.7 & \textbf{17.3} \\
                                & 7          & 0.74 & 0.77 & 21.1 & 15.6 \\
                                & \textbf{8} & \textbf{0.68} & 0.78 & 21.8 & 16.2 \\
                                & 9          & 0.69 & \textbf{0.79} & \textbf{22.3} & \textbf{17.3} \\

\hline

\multirow{6}{2.5cm}{Basal Cell \newline Carcinoma } & 4           & 0.47 & 0.71 & 20.9 & 17.2 \\
                                                   & 5           & 0.42  & 0.70 & 20.0 & 17.2 \\
                                                   & \textbf{6}  & 0.39  & 0.69 & 20.3 & 17.4 \\
                                                   & 7           & \textbf{0.37} & 0.70 & 21.1 & 15.6 \\
                                                   & 8           & 0.41 & 0.74 & 23.2 & 17.6 \\
                                                   & 9           & 0.41 & \textbf{0.75} & \textbf{24.2} & \textbf{19.0} \\
\hline

\multirow{6}{2.5cm}{Benign \newline Keratosis} & 4           & 0.46 & 0.84 & 39.0 & 31.0 \\
                                              & 5           & 0.48 & 0.85 & 40.4 & 33.4 \\
                                              & 6           & 0.42 & 0.85 & 40.2 & 33.7 \\
                                              & \textbf{7}  & 0.43 & 0.85 & 41.0 & 29.7 \\
                                              & 8           & \textbf{0.41} & 0.86 & 42.9 & 32.2\\
                                              & 9           & 0.43 & \textbf{0.87} & \textbf{44.8} & \textbf{34.3}\\
\hline

\multirow{6}{2.5cm}{Dermatofibroma} &4            & 0.62 & 0.53 & 10.0 & 5.34 \\
                                   & 5           & 0.70 & 0.55 & 11.2 & 6.31 \\
                                   & 6           & 0.72 & 0.57 & 11.7 & \textbf{7.75}\\
                                   & \textbf{7}  & \textbf{0.57} & 0.57 & 11.7 & 4.62 \\
                                   & 8           & 0.59 & \textbf{0.60} & 12.2 & 5.44\\
                                   & 9           & 0.61 & \textbf{0.60} & \textbf{13.2} & 5.60 \\
\hline

\multirow{6}{2.5cm}{Vascular \newline Lesions} & 4           & 0.68 & 0.62 & 19.1 & 13.4 \\
                                              & 5           & 0.60  & 0.62 & 19.1 & 13.4 \\
                                              & \textbf{6}  & 0.62  & 0.64 & 20.7 & 15.2 \\
                                              & 7           & \textbf{0.57}  & 0.64 & 20.8 & 11.9\\
                                              & 8           & 0.65  & 0.68 & 24.0 & 16.1 \\
                                              & 9           & 0.62  & \textbf{0.70} & \textbf{24.7} & \textbf{16.8}\\
\hline

\multirow{6}{2.5cm}{Squamous Cell \newline Carcinoma}  & 4           & 0.43 & 0.40 & 3.55 & 2.09\\
                                                      & \textbf{5}  & 0.39 & 0.41 & 3.96 & 2.50\\
                                                      & 6           & 0.46 & 0.47 & 4.98 & \textbf{3.22}\\
                                                      & 7           & 0.42 & 0.48 & 5.58 & 2.62\\
                                                      & 8           & 0.43 & \textbf{0.49} & \textbf{5.88} & 2.94\\
                                                      & 9           & \textbf{0.37} & 0.45 & 5.17 & 2.70\\
\hline

\end{tabular}
\end{table*}

\section{Ablation Study on the scaling parameter $\alpha$}

To evaluate the sensitivity of the scaling parameter $\alpha$ for the detected concepts, experiments are conducted across representative concepts. Table~\ref{tab:ablation} reports the effect of varying $\alpha$ on three metrics: Success Rate (SR), which indicates the proportion of test samples for which traversal along the corresponding direction increases the predicted probability of the target class relative to the reconstructed image, LPIPS~\cite{zhang2018unreasonable} measures perceptual realism of generated images (lower is better), and FID~\cite{heusel2017gans} assesses overall image distributional fidelity (lower is better). 

The results reveal a trade-off: smaller $\alpha$ values yield more realistic images (lower LPIPS/FID) but weaker concept manifestation (lower SR), whereas larger $\alpha$ values improve SR but may introduce artifacts and reduce realism. The optimal $\alpha$ for each concept is highlighted in bold and reflects a balanced trade-off between edit success and perceptual fidelity.  

Notably, the best-performing $\alpha$ varies across concepts (e.g., $40$ for \textit{Purplish Core Pigmentation}, and $70$ for \textit{Reddish Core with Blue-Gray Dots}), confirming that a single global value of $\alpha$ is suboptimal. For most concepts, moderate $\alpha$ values provide a stable regime where SR increases steadily with only minor degradation in perceptual metrics, whereas higher values often exaggerate the edits, leading to artifacts despite marginal SR gains.

\begin{table}[!h]
\caption{\label{tab:ablation} Ablation study on the effect of $\alpha$ for selected concept directions. The best values for each metric are highlighted in bold. The chosen $\alpha$ (bold entry in the first column) represents the optimal balance between concept success rate and perceptual fidelity. Lower LPIPS and FID values indicate higher perceptual quality and image realism.\\} 
\centering
\renewcommand{\arraystretch}{1.3}
\begin{tabular}{L{2.0cm} C{6mm} C{18mm} C{12mm} C{8mm}}
\hline
\textbf{Concept} & $\boldsymbol{\alpha}$ & \textbf{SR (\%)}~$\uparrow$ & \textbf{LPIPS}~$\downarrow$ & \textbf{FID}~$\downarrow$ \\
\hline

\multirow{5}{2.2cm}{\centering Purplish Core \\ Pigmentation} 
 & \textbf{40} & \textbf{70.54} & \textbf{0.123} & \textbf{30.51}\\
 & 45 & 69.86 & 0.130 & 35.56\\
 & 50 & 68.34 & 0.136 & 40.42\\
 & 55 & 66.23 & 0.143 & 45.08\\
 & 60 & 63.45 & 0.150 & 49.77\\
\hline

\multirow{5}{2.2cm}{\centering Reddish \\ Core with \\ Blue-Gray Dots}
 & 50 & 74.68 & \textbf{0.138} & \textbf{22.11}\\
 & 55 & 77.94 & 0.145 & 26.58\\
 & 60 & 80.31 & 0.153 & 31.73\\
 & 65 & 81.84 & 0.160 & 37.10\\
 & \textbf{70} & \textbf{83.61} & 0.168 & 43.05\\
\hline

\multirow{5}{2.2cm}{\centering Blotchy\\ Pigmentation with\\ Irregular Texture}
 & 70 & 72.61 & \textbf{0.165} & \textbf{29.10}\\
 & 75 & 75.36 & 0.173 & 33.12\\
 & 80 & 78.11 & 0.181 & 37.65\\
 & 85 & 79.50 & 0.189 & 42.13\\
 & \textbf{90} & \textbf{81.09} & 0.198 & 47.01\\ 
\hline

\multirow{5}{2.2cm}{\centering Central Pinkish\\ Veil with Asymmetric \\ Intensified Pigmentation}
 & 40 & 80.28 & \textbf{0.124} & \textbf{32.04}\\
 & 45 & 82.35 & 0.131 & 35.86\\
 & 50 & 83.88 & 0.137 & 39.54\\
 & 55 & 84.96 & 0.143 & 43.29\\
 & \textbf{60} & \textbf{85.44} & 0.149 & 47.13\\
\hline

\multirow{5}{2.2cm}{\centering Central \\ Purplish Veil}
 & 40 & 76.55 & \textbf{0.12}5 & \textbf{34.61}\\
 & 45 & 79.13 & 0.132 & 39.63\\
 & 50 & 81.26 & 0.138 & 44.31\\
 & 55 & 82.69 & 0.145 & 48.90\\
 & \textbf{60} & \textbf{84.59} & 0.152 & 53.48\\
\hline

\multirow{5}{2.2cm}{\centering White Structures\\ with Irregular Vessels}
 & 60 & 51.54 & \textbf{0.148} & \textbf{31.35}\\
 & 65 & 56.83 & 0.155 & 36.14\\
 & 70 & 62.50 & 0.163 & 40.98\\
 & 75 & 66.30 & 0.170 & 45.98\\
 & \textbf{80} & \textbf{70.47} & 0.177 & 51.21\\

\hline
\end{tabular}
\end{table}

\section{Evaluation of concept directions: TCAV score}

\begin{table*}[!h]
\caption{\label{tab:tcav} TCAV scores for representative concepts across different layers of the ResNet50 architecture. Higher scores indicate stronger alignment between the concept direction and the model’s decision boundary. Bold entries denote the layer with the highest score for each concept. Here, layer4-B0 (conv3), layer4-B1 (conv3), and layer4-B2 (conv3) refer to the final conv3 outputs of the first, second, and third bottleneck blocks in the layer 4, respectively. \\} 
\centering
\renewcommand{\arraystretch}{1.3}
\begin{tabular}{L{3.5cm} C{2cm} C{2cm} C{2cm} C{2cm} C{2cm} C{2cm}}
\hline
\textbf{Concept} & \textbf{layer1} & \textbf{layer2} &  \textbf{layer3} & \textbf{layer4-B0 (conv3)} & \textbf{layer4-B1 (conv3)} & \textbf{layer4-B2 (conv3)}\\
\hline

\makecell[l]{Purplish Core\\ Pigmentation} & $0.46\pm0.04$ & $0.35\pm0.06$ & $0.31\pm0.13$ & $0.39\pm0.10$ & $0.77\pm0.02$ & $\mathbf{0.97\pm0.02}$\\
\hline

\makecell[l]{Reddish Core with \\ Blue-Gray Dots}
& $0.73\pm0.01$ & $\mathbf{0.93\pm0.01}$ & $0.89\pm0.03$ & $0.77\pm0.07$ & $0.82\pm0.07$ & $0.82\pm0.13$\\

\hline

\makecell[l]{Blotchy Pigmentation\\ with Irregular Texture}
 & $0.64\pm0.02$ & $0.76\pm0.01$ & $0.73\pm0.10$ & $0.72\pm0.08$ & $0.86\pm0.07$ & $\mathbf{0.92\pm0.02}$\\
 
\hline

\makecell[l]{Central Pinkish Veil\\ with Asymmetric\\ Intensified Pigmentation}
 & $0.70\pm0.01$ & $0.96\pm0.01$ & $0.88\pm0.01$ & $0.95\pm0.02$ & $\mathbf{1.00\pm0.00}$& $\mathbf{1.00\pm0.00}$\\
 
\hline

\makecell[l]{Central Purplish Veil}
 & $0.72\pm0.02$ & $0.77\pm0.02$ & $0.96\pm0.02$ & $0.98\pm0.01$ & $0.98\pm0.00$ & $\mathbf{1.00\pm0.00}$\\

\hline

\makecell[l]{White Structures\\ with Irregular Vessels}
 & $\mathbf{0.76\pm0.01}$ & $0.66\pm0.00$ & $0.65\pm0.00$ & $0.66\pm0.06$ & $0.74\pm0.10$ & $0.64\pm0.18$\\

\hline
\end{tabular}
\end{table*}

Table~\ref{tab:tcav} reports the TCAV~\cite{kim2018interpretability} scores computed across different layers of  ResNet50~\cite{he2016deep} (target model) architecture for representative concepts. Higher TCAV values indicate a stronger alignment between the concept direction and the model’s decision boundary. 

Several insights emerge from these results. First, concept importance varies substantially across layers, highlighting that semantic information is not uniformly distributed throughout the network. For example, \textit{Purplish Core Pigmentation} exhibits a clear progression from low scores in early layers (0.46 in layer1) to a very strong alignment in deeper layers (0.97 at the final conv3 of the second bottleneck in layer4), suggesting that this concept is primarily captured at higher semantic levels. Similarly, \textit{Blotchy Pigmentation with Irregular Texture} and \textit{Central Purplish Veil} both achieve their peak TCAV scores in deeper blocks, indicating that these nuanced features emerge late in the representation hierarchy. 

In contrast, \textit{Reddish Core with Blue-Gray Dots} reaches its maximum importance at an intermediate layer (0.93 in layer2), suggesting that certain concepts are best represented at mid-level feature abstraction. Interestingly, \textit{White Structures with Irregular Vessels} shows the opposite trend, with its highest score at the shallowest layer (0.76 in layer1), implying that this concept is captured through low-level visual cues such as edge or color contrasts. 

These findings demonstrate that different clinical concepts are encoded at different depths of the network: some emerge only in higher-level semantic layers, while others are already salient at earlier stages. This underlines the importance of evaluating multiple layers when applying concept-based interpretability methods.

\section{Extraction of Concept Directions Using the CDCT Encoder}

This supplementary section presents the experimental setup used to extract concept directions using the encoder from the CDCT framework (hereafter referred to as the CDCT encoder).
The CDCT encoder is a Variational Autoencoder (VAE) trained on a counterfactual trajectory dataset generated via a latent diffusion model with classifier guidance, as described in Step 2 (\textit{Semantic Space Disentanglement}) of the CDCT framework~\cite{varshney2025generating}.
The training objective encourages the latent space to encode distinct, classifier-relevant transformations, facilitating more interpretable and disentangled concept representations. Further architectural and training details are provided in the original work~\cite{varshney2025generating}.

In this supplementary analysis, the same pipeline described in \textit{Section 3.2} of the main manuscript is applied to compute latent difference vectors and extract concept directions.
Specifically, for each training sample not predicted as the target class, a classifier-guided counterfactual is generated, and both the factual and counterfactual images are encoded using the CDCT encoder.
Unlike the LDM encoder, which yields latent embeddings of shape $4 \times 32 \times 32$, the CDCT encoder produces compact latent representations of shape $512 \times 1 \times 1$.
Each latent difference is flattened into a 512-dimensional vector and normalized to unit length to produce directional vectors.
Spherical K-Means clustering~\cite{hornik2012spherical} is then applied to the set of unit-norm latent differences computed per class.
The number of clusters is selected based on the silhouette coefficient to capture cohesive and distinct semantic directions. 
The average unit direction within each resulting cluster is interpreted as a representative concept direction for the target class.

The results highlight how concept representations vary in the CDCT encoder’s latent space, particularly in terms of semantic disentanglement and reconstruction fidelity.
Visualizations and class-specific examples are provided in the following subsections to support a comparative evaluation between the CDCT and LDM encoders.

\section{Interpretation and Evaluation of Concept Directions Using the CDCT Encoder}
\begin{figure*}[htp]
    \centering  
    \includegraphics[width=.75\textwidth]{Column_Label.jpg}
    \centering
    \subfloat[Success Rate = 65.62 \% - Peripheral Darkening and Border Irregularity.]{\label{fig:MEL_CDCT}
    \includegraphics[width=.75\textwidth]{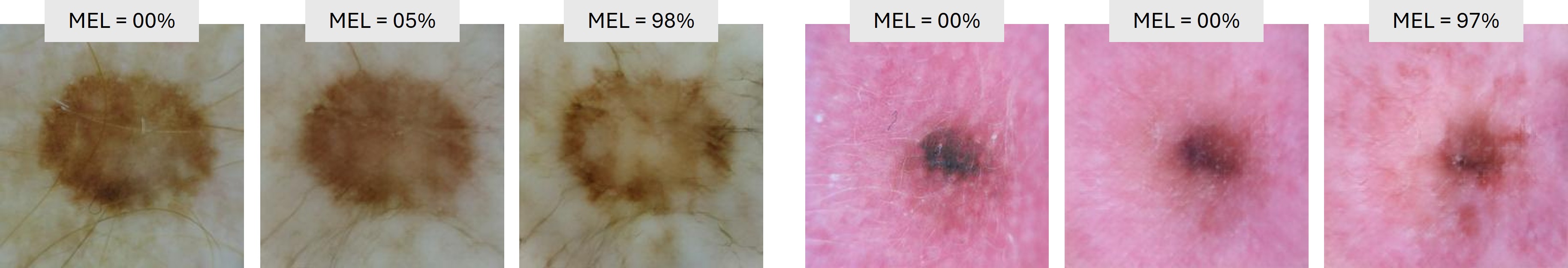}} \\    
    \subfloat[Success Rate = 77.53 \% - De-pigmentation and Soft-Border.]{\label{fig:BCC_CDCT}
      \includegraphics[width=.75\textwidth]{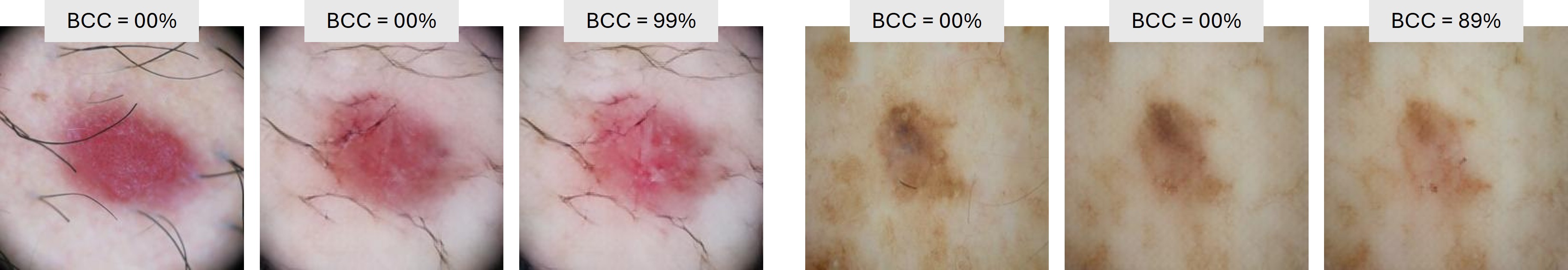}} \\
    \subfloat[Success Rate = 65.21 \% - Scaly, Lightened Center with Red Halo.]{\label{fig:AK_CDCT}
      \includegraphics[width=.75\textwidth]{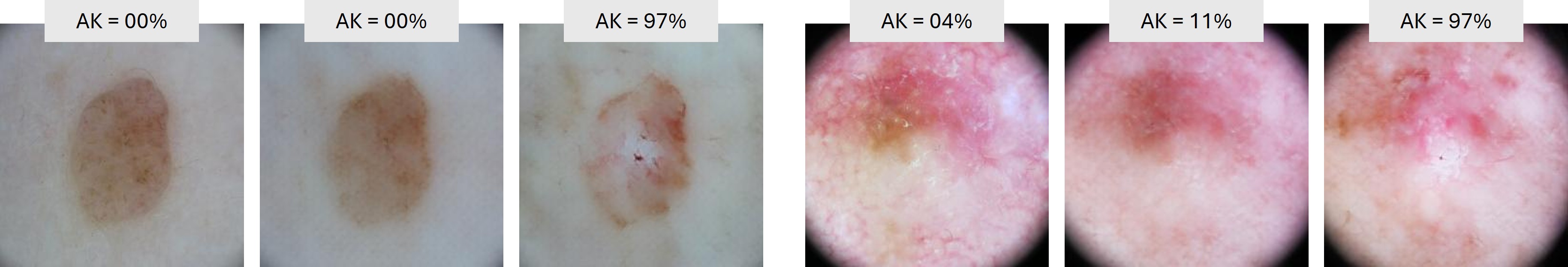}} \\    
    \subfloat[Success Rate = 70.03 \% - Reddish Center with Surrounding Pigmentation.]{\label{fig:BKL_CDCT}
      \includegraphics[width=.75\textwidth]{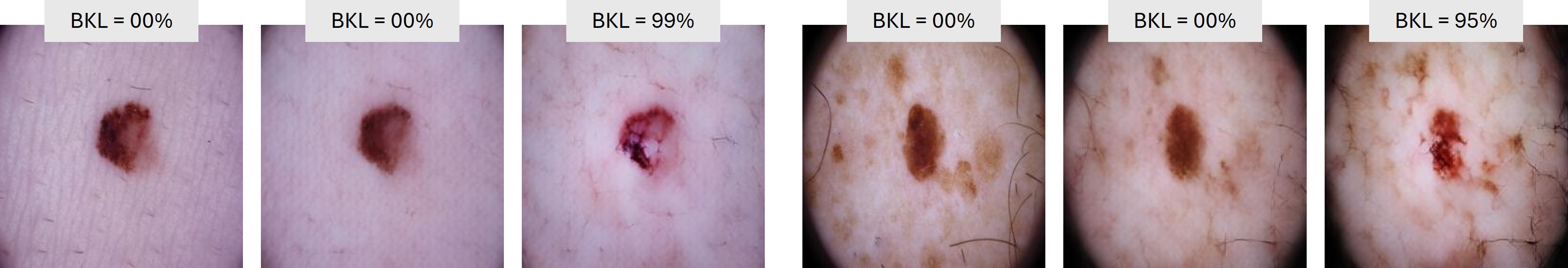}} \\      
    \subfloat[Success Rate = 76.17 \% - Central Scar-Like Area with Pink Hue.]{\label{fig:DF_CDCT}
      \includegraphics[width=.75\textwidth]{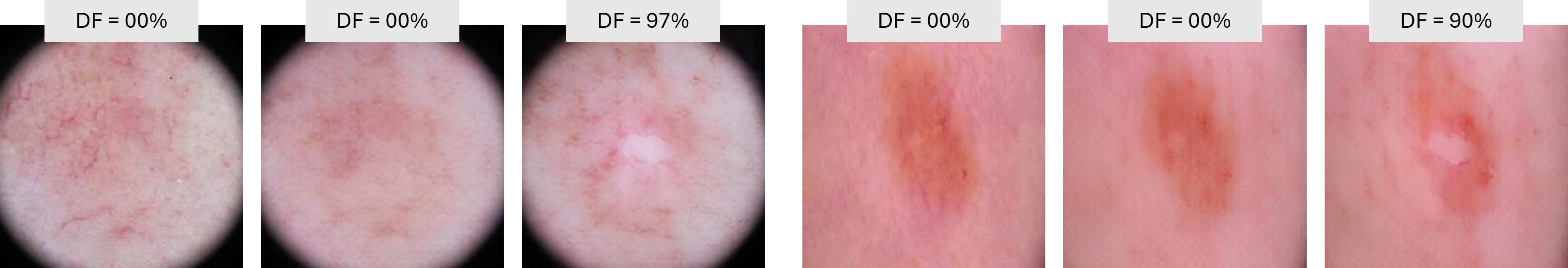}} \\      
    \subfloat[Success Rate =  64.50 \% - Central Redness.]{\label{fig:VASC_CDCT}
      \includegraphics[width=.75\textwidth]{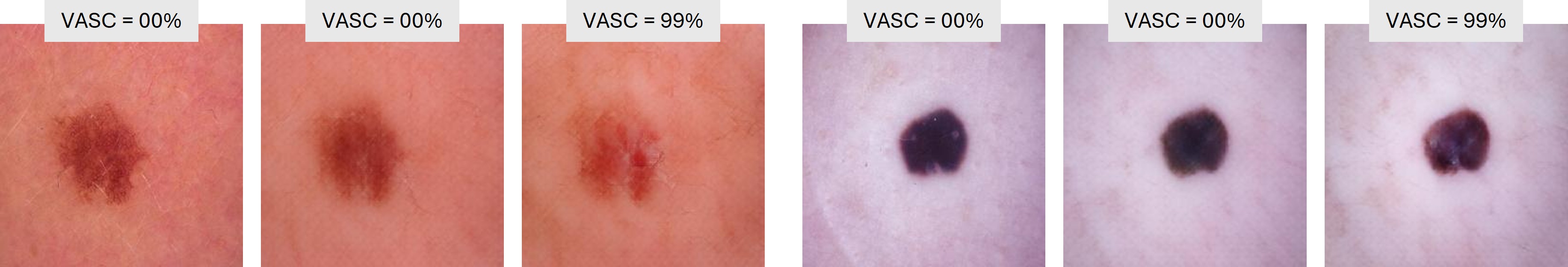}} \\      
    \subfloat[Success Rate = 85.64 \% - White Center with Red Surround.]{\label{fig:SCC_CDCT}
      \includegraphics[width=.75\textwidth]{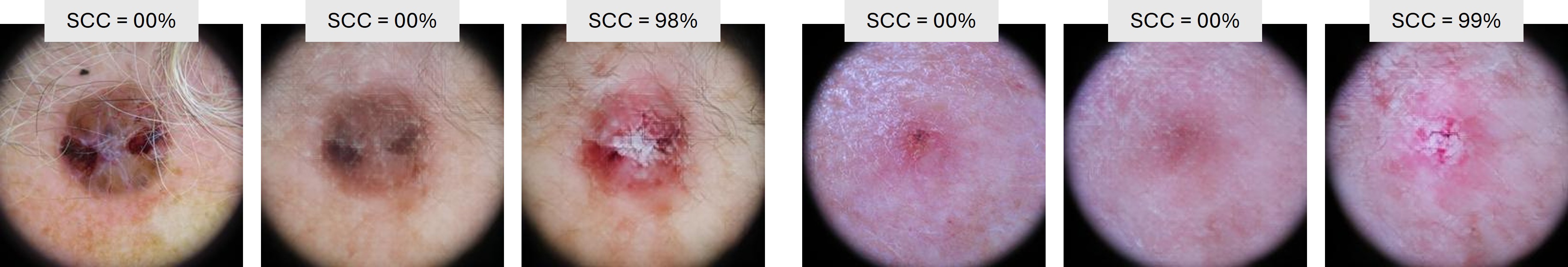}}      
    \caption{Discovered concepts by \frameworkAcronym\ on the ISIC dataset using the CDCT encoder. Each row shows two examples: original, reconstructed, and manipulated images (left to right). The predicted probability for the target class associated with each concept direction is shown above each image.}
    \label{fig:concepts_CDCT}
\end{figure*}

To evaluate the semantic relevance and discriminative influence of the discovered concept directions, each direction identified by the CDCT encoder is applied to the latent representation of previously unseen test samples.
Figure~\ref{fig:concepts_CDCT} presents representative concepts produced using the CDCT encoder.
For each direction, the accompanying success rate (reported in each sub-caption) quantifies the proportion of test cases in which latent traversal increases the predicted probability of the corresponding target class, relative to the classifier’s output on the reconstructed factual image.

The concept directions extracted using the CDCT encoder exhibit class-specific semantic transformations aligned with known dermoscopic features.
For \textit{Melanoma}, the direction (Figure~\ref{fig:MEL_CDCT}) darkens the lesion periphery, reduces central contrast, and softens the borders.
The \textit{Basal Cell Carcinoma} direction (Figure~\ref{fig:BCC_CDCT}) reduces pigmentation and softens edge contrast, yielding a more blended lesion appearance.
In the case of \textit{Actinic Keratosis} (Figure~\ref{fig:AK_CDCT}), the identified direction introduces a central lightened area with a scaly texture and peripheral reddish hue.
The \textit{Benign Keratosis} direction (Figure~\ref{fig:BKL_CDCT}) induces reddish central pigmentation with uneven borders and surrounding pigment variation.
For \textit{Dermatofibroma}, the concept direction (Figure~\ref{fig:DF_CDCT}) enhances a pale central region surrounded by a subtle pink hue, features reported in literature~\cite{agero2006conventional}.
The direction for \textit{Vascular Lesion} (Figure~\ref{fig:VASC_CDCT}) reveals central redness accompanied and subtle irregularities in pigment distribution, resembling vascular blushes typically observed in angiomas and angiokeratomas~\cite{ piccolo2018dermatoscopy}
Finally, the \textit{Squamous Cell Carcinoma} direction (Figure~\ref{fig:SCC_CDCT}) introduces a central white, keratin-like structure surrounded by inflamed reddish tissue, consistent with dermoscopic features described by Rosendahl et al.~\cite{rosendahl2012dermoscopy}.

While several concept directions extracted using the CDCT encoder correspond to clinically relevant dermoscopic patterns, such as keratin-like centers in \textit{Squamous Cell Carcinoma} or pale regions in \textit{Dermatofibroma}, these findings are less detailed visually than those produced by the LDM encoder. 
The CDCT encoder, despite being explicitly trained to disentangle classifier-relevant transformations, demonstrates lower success rates across most classes in comparison to results produced by the LDM encoder.
This can be attributed to its limited ability to reconstruct fine-grained dermoscopic features, which are critical for clinical interpretation.
Furthermore, its lower-dimensional latent space restricts its capacity to model complex transformations, further limiting its effectiveness in capturing meaningful semantic changes.

Although many of the extracted directions align with features documented in dermatological literature, their diagnostic relevance remains to be clinically validated.
Expert review by dermatologists is essential to determine whether these directions reflect genuine clinical cues, dataset-specific artifacts, or potentially novel biomarkers.
Such validation would enhance the interpretability and utility of concept-based explanations in high-stakes medical applications.

\clearpage

\bibliographystyle{elsarticle-num} 
\bibliography{main}

\end{document}